\documentclass[cameraready]{Interspeech}

\title{Flexible and Interpretable Accent Distance Measurements\thanks{This research is funded by an EPSRC DTP and the Vice-Chancellor's Award. It is supported by Cambridge University Press \& Assessment, a department of The Chancellor, Masters, and Scholars of the University of Cambridge.}}

\author[affiliation={1}]{Charles}{McGhee}
\author[affiliation={1}]{Mark J. F.}{Gales}
\author[affiliation={1}]{Kate M.}{Knill}

\address{
    $^1$ ALTA Institute/MIL, Department of Engineering, University of Cambridge, UK\
}

\email{cgm43@cam.ac.uk, mjfg@eng.cam.ac.uk, kmk1001@cam.ac.uk}


\usepackage{comment}
\usepackage{multirow}
\usepackage{mathtools}
\usepackage{tipa}
\usepackage{float}

\begin{document}

\maketitle

\begin{abstract}
    Determining the differences between two speakers' accents is a fundamental task in linguistics and speech technology research. The methodology used to measure these differences depends on the specific research area. A phonetics researcher may demonstrate accent variation by comparing vowel formants in paired recordings of individual words. These results will be interpretable, but the recordings will be time-consuming to collect and may not be representative of connected speech. Accented Text-to-Speech (TTS) research has pushed towards using accent embeddings derived from accent classification tasks. These embeddings can be produced from any speech recording, but are not readily interpretable. In this paper, we demonstrate that articulatory representations created through articulatory inversion can be used as an interpretable basis for accent comparison and that optimal transport provides a framework for accent comparison across arbitrary recording types.
\end{abstract}

\section{Introduction}

An accent can be defined as the segmental and suprasegmental elements of pronunciation that allow a listener to identify where a speaker might be from in terms of region or social class \cite{crystal2011dictionary,markl2023everyone}.
Being able to take utterances from two speakers and measure specific and global differences in their pronunciation is a core problem in linguistics and speech technology research.
However, the methodology used by different researchers can vary according to the field.

For a researcher working in linguistics, a common goal is to measure regional differences in pronunciation and demonstrate how this fits in with the historical context of the region and within a broader phonological framework.
The differences in pronunciation need to be interpretable, so a typical experiment might be to measure the difference between speaker-normalised formants across vowels in minimal sets, e.g. `hood', `hid', `had' \cite{ferragne2010vowel}. 
Collecting enough recordings to make meaningful conclusions is time-consuming, and the simple structure of samples may not reflect the reality of pronunciation in connected or conversational speech.
It would therefore be useful to have a method that can work for a range of phonetic contexts whilst still being interpretable enough to determine specific phonetic differences between two speakers.

From a machine learning perspective, we often want a metric that tells us `at-a-glance' how different two speakers are in terms of accent.
Recent work has focused on the identity element of accent, as embedding models like the Generalisable Accent Identification (GenAID) model \cite{zhong2025accentbox} are trained to perform utterance-wise classification of national-level accent labels on datasets like CommonAccent \cite{zuluagagomez23_interspeech}.
We expect the model to capture a more holistic definition of an accent, due to its training criteria and the broadness of the label used.
Humans use a range of cues to identify different accents \cite{derwing2009putting}, but both these embeddings and human cognitive processes are less readily analysable than the vowel-distance method mentioned above.

In this work, we attempt to fill the gap between the two approaches using deep articulatory representations \cite{mcghee25_interspeech} and Optimal Transport (OT) \cite{peyre2019computational}.
The deep articulatory basis for pronunciation comparison allows us to identify specific phonetic differences (e.g. rhoticity), whilst remaining robust to speaker identity and recording conditions.
Optimal transport then allows us to move away from paired recordings to utterance-independent accent comparison, in a manner similar to an accent embedding.
We begin with a description of the articulatory feature set and optimal transport before presenting our results comparing speakers from the Voice Cloning Toolkit (VCTK) \cite{yamagishi2019cstr}.

\section{Background}

\subsection{Deep Articulatory Representations}

Acoustic-to-Articulatory Inversion (AAI) is the process of estimating articulation from speech. 
Articulation can be measured using fleshpoint sensor measurements, as in X-Ray Microbeam (XRMB), or through imaging techniques like ultrasound or Magnetic Resonance Imaging (MRI).
The sensor positions or segmented images are used as regression targets and estimated from speech, typically represented using some form of Self-Supervised Learning (SSL) speech model, such as WavLM \cite{chen2022wavlm}.
A complete AAI model would be able to recover a specific speaker's articulation, including any differences in vocal tract morphology.
However, for pronunciation assessment, we would prefer that the output articulation is the same for speech sounds from the same phonetic category across different speakers.
In \cite{mcghee25_interspeech}, an AAI model trained on the Wisconsin XRMB (WXRMB) dataset \cite{westbury1994x} was adapted to reduce the distance between articulatory targets for similar sounds between several speakers.
These speaker-consistent articulatory representations were expanded upon in \cite{mcghee2025_slate} to include a 2D PCA representation derived from WavLM that captured voicing and nasality, known as the Voicing, Vowels and Nasality (VVN) feature.
The full representation contains 6 2D points from WXRMB representing the tongue (T3-T1), lips (UL and LL) and jaw (MANi), and the 2D VVN feature, giving a 14-dimensional representation of speech which we refer to as Art+VVN throughout.
We will use this feature set, as well as final-layer WavLM features, as the basis for optimal transport and other comparison methods introduced in Section 3.

\subsection{Optimal Transport}

There are many situations in which it is either impractical or impossible to collect paired speech data.
We would therefore like to be able to take sets of unpaired utterances from different speakers and compare them.
Optimal transport provides a framework to compare two unstructured distributions and determine the cost of transforming one into the other.
We use the notation from  \cite{peyre2019computational} throughout. 
For two speakers with acoustic features of dimension $f$,  $\textbf{X} \in \mathbb{R}^{n\times f}$ and $\textbf{Y} \in \mathbb{R}^{m\times f}$, the cost matrix $\textbf{C} \in \mathbb{R}^{n\times m}$ using a Euclidean cost is defined as:

\begin{equation}
    \textbf{C}_{i,j} = ||\textbf{x}_{i} -\textbf{y}_{j}||_{2}
\end{equation}

Each point has a weighting forming distributions $\textbf{a} \in \Sigma_{n}$ and $\textbf{b} \in \Sigma_{m}$ where $\Sigma$ is the probability simplex:

\begin{equation}
    \Sigma_{n} = \left\{ \textbf{a} \in \mathbb{R}^{n}_{+} : \sum^{n}_{i=1} \textbf{a}_{i} = 1 \right\}
\end{equation}

In the general case, where $n$ and $m$ can assume different values, we can state the optimal transport problem as finding the optimal coupling matrix $\textbf{P}$, where $\textbf{P}_{i,j}$ gives the flow of mass from bin $i$ to bin $j$ and minimises the following cost:

\begin{equation}
\label{original_cost}
    L_{\textbf{C}}(\textbf{a},\textbf{b}) = \min_{\textbf{P} \in \textbf{U}(\textbf{a},\textbf{b})} \sum_{i,j} \textbf{C}_{i,j}\textbf{P}_{i,j}
\end{equation}

$\textbf{U}(\textbf{a},\textbf{b})$ is the set of coupling matrices defined:

\begin{equation}
\label{equality constraints}
    \textbf{U}(\textbf{a},\textbf{b}) = \left\{ \textbf{P} \in \mathbb{R}^{n\times m}_{+} : \sum_{i} \textbf{P}_{i,j} = \textbf{a},    \sum_{j} \textbf{P}_{i,j} = \textbf{b} \right\}
\end{equation}

For our application, we ensure $n=m$ and assign a uniform weighting to each point.
Under these assumptions, the coupling matrix $\textbf{P}$ is a permutation matrix describing the point-to-point movement of each feature and has an optimal solution $\textbf{P}^*$ that can be found using the network simplex algorithm (see Chapter 3 in \cite{peyre2019computational}).
We take the cost $L_{\textbf{C}}(\textbf{a},\textbf{b})$ associated with the optimal solution $\textbf{P}^*$ as the distance metric between the feature spaces of the two speakers.

\section{Evaluation}

Our definition of accent in the introduction consisted of those elements of pronunciation that allowed ready identification of region or social class.
To this end, we evaluate each distance measure across two major axes, namely identity and phonetic distance.

\subsection{Identity}

By identity, we mean both accent and speaker identity,
We go by the assumption that two speakers with the same accent can have very different speaker characteristics (although there is likely some overlap \cite{schwab2016speakers}) and therefore would like a distance metric that groups similar accents together, but not similar speakers.
To measure this, we will use both an accent classification task and also see how well the proposed distance measurements correlate with speaker embedding distances from an off-the-shelf speaker verification model\footnote{https://huggingface.co/microsoft/wavlm-base-plus-sv}.
We use Spearman Rank Correlation (SRC) between the distance from each method and the distances measured between speaker embeddings for a given speaker against all other speakers.
We then average the SRC values across the whole speaker set.
This same method is used for all correlation calculations, including those between phonetic distance metrics introduced in the next section.

For accent classification, we will look at three-way national-level labels for Scottish, English (England) and Irish (merging Irish and Northern Irish, as in CommonAccent \cite{zuluagagomez23_interspeech}) speakers from VCTK.
We use this dataset as it contains a large amount of data per speaker, paired/unpaired recordings and some regional information for each speaker that we use later when considering how each method views the relation between each region.
These three regions are selected as there is existing literature on the general differences/similarities between the regions \cite{ferragne2010vowel} and GenAID was trained to predict these three broad classes \cite{zhong2025accentbox}.
We classify a speaker as one of the accents by comparing that speaker against every other speaker in the set and choosing the class that has the lowest average distance to the candidate speaker.
This method allows us to retain the same classification methodology across the phonetic distances established in the next section and embeddings derived from GenAID and the speaker verification model ('Spk Emb' in results).
For both embedding models, we average the utterance-level embeddings across each speaker to create a speaker-level embedding.

\subsection{Phonetic Distance}

For phonetic distance, we look at measuring the differences between speakers across vowels, utterances and sets of utterances.
We want to see how well the distance metrics correlate with each other and how these results compare with accent classification and speaker distance results.
This should allow us to elicit any differences in methods, e.g. between pronunciation and embedding distances.
We will look first at vowel distances, then at aligning the features between paired utterances and finally OT across arbitrary utterance types.

\subsubsection{Vowel Distances}

By isolating specific vowel distances, we ignore the suprasegmental properties of speech and any consonantal differences.
This gives us a baseline from which we can compare alignment and OT.
We use the Montreal Forced Aligner (MFA) to find vowel positions within an utterance and then take the midpoint for three different feature sets: our articulatory features as above, final-layer WavLM features and formants extracted using the method from \cite{shrem22_interspeech}.
For the articulatory features and formants, we use Euclidean distance, and for the WavLM features, we use cosine distance.
Being an automatic method, it is susceptible to alignment errors and does not allow us to use the standard scaling methods for vowel tokens in CVC formats (we simply z-score F1 and F2 for each speaker).

\subsubsection{Alignment}

In \cite{mcghee2025_slate}, articulatory features were used to align L2 speakers of English with template productions, and it was found that the utterance-level alignment error strongly correlated (r=0.88) with human assessments of utterance-level pronunciation error.
For the present study, we use a slightly different DTW step pattern (combination IB from Sec 4.7 in \cite{rabiner1993fundamentals}), as we found that the standard implementation in the \verb|dtw-python| \footnote{https://pypi.org/project/dtw-python/} package can often skip over sections with very different pronunciation and therefore not capture errorful sections correctly. 
In comparing alignment, we use articulatory features and final-layer WavLM features with Euclidean and cosine distances, respectively.

\subsubsection{Optimal Transport}

We use the package \verb|POT| \footnote{https://pypi.org/project/POT/} package to implement optimal transport.
As above, this package uses the network simplex algorithm to compute exact optimal transport with a time complexity of $\mathcal{O}(n^3)$.
Our articulatory/WavLM features are output at 50Hz and the speakers from VCTK have $\sim$12.5 mins of speech recorded each (ignoring silences), so the feature set for each speaker is too large to perform OT on directly.
We opt to reduce the number of input points to OT by performing k-means clustering (k=1000) on the whole feature set and using the cluster centres as input to OT.
Features are extracted for segments with speech only, discarding silence according to MFA timestamps.
In section 4.4, we show that the amount of input data and the number of clusters can have a significant impact on the performance of this method.

\section{Results}

\subsection{Vowel Distances and Alignment}

We first present a comparison of the more standard measures of phonetic distance (vowel distances and DTW feature alignment) with GenAID and the verification model-derived speaker embedding (Spk Emb) in terms of classification performance and speaker embedding correlation.
In Table \ref{src_class}, we show that alignment with articulatory features can achieve the same classification results as GenAID embeddings, whilst simultaneously having lower correlation with speaker attributes. Note, all methods failed on a specific speaker (p285) who appears to be mislabelled.

\begin{table}[h]
\caption{Accent classification accuracy and SRC with speaker embedding for different approaches.}
\label{src_class}
\centering
\begin{tabular}{cccc}
\hline
\multicolumn{2}{c}{\textbf{Method}}                  & \textbf{Spk Emb} & \textbf{\% Class. Acc} \\ \hline
\multirow{3}{*}{\textbf{\begin{tabular}[c]{@{}c@{}}Vowel \\ Distance\end{tabular}}} & \textbf{Formants}  & 0.28             & 74                    \\
                                & \textbf{WavLM}     & 0.22             & 88                    \\
                                & \textbf{Art + VVN} & 0.12    & 91                    \\ \hline
\multirow{2}{*}{\textbf{Alignment}} & \textbf{WavLM}     & 0.28             & 94                    \\
                                & \textbf{Art + VVN} & 0.17             & 99           \\ \hline
\multirow{2}{*}{\textbf{Embedding}}                       & \textbf{GenAID}    & 0.29             & 99           \\
\textbf{}                       & \textbf{Spk Emb}   & \textbf{-}       & 67                    \\ \hline
\end{tabular}
\end{table}

\begin{table*}[!t]
\caption{Accent classification accuracy and SRC with methods from Table \ref{src_class} for optimal transport and GenAID.}
\label{big_table}
\centering
\label{tab:my_big_table}
\begin{tabular}{ccccccccc}
\cline{3-7}
                                                                                       &                    & \multicolumn{2}{c}{\textbf{Alignment}} & \multicolumn{3}{c}{\textbf{Vowel Distance}}            &                  &                       \\ \hline
\multicolumn{2}{c}{\textbf{Method}}                                                                         & \textbf{WavLM}   & \textbf{Art + VVN}  & \textbf{WavLM} & \textbf{Art +VVN} & \textbf{Formants} & \textbf{Spk Emb} & \textbf{\% Class Acc} \\ \hline
\multirow{2}{*}{\textbf{\begin{tabular}[c]{@{}c@{}}Optimal \\ Transport\end{tabular}}} & \textbf{WavLM}     & 0.80    & 0.59                & 0.69           & 0.58              & 0.42     & 0.39             & 96                    \\
                                                                                       & \textbf{Art + VVN} & 0.65             & 0.81       & 0.70  & 0.77     & 0.39              & 0.16    & 94                    \\ \hline
\textbf{Embedding}                                                                     & \textbf{GenAID}    & 0.52             & 0.59                & 0.55           & 0.55              & 0.32              & 0.29             & 99           \\ \hline
\end{tabular}
\end{table*}

DTW aligning the features from each speaker outperforms the vowel distance metrics, likely capturing accent-relevant consonantal and suprasegmental details that are lost by only taking vowel distances.
Final-layer WavLM features also have higher correlations with speaker embeddings than our articulatory features.
This is expected, as we train our inversion model to ignore speaker information and the 14-dimensional articulatory representation contains far less information than the full 1024-dimensional WavLM Large feature.

One key feature that is missing in the articulatory representation is F0, but the classification results with alignment here would indicate that this is less relevant for classifying these specific accents at such a broad level.
We also note that the speaker embedding had generally quite poor correlation across all other accent distance measurements, but still retained reasonably high classification accuracy.
As we mentioned in Section 3, we assume that accent and speaker identity are mostly disentangled, but some global speaker properties may help to predict accent identity \cite{schwab2016speakers,derwing2009putting}.

\subsection{Optimal Transport}

In Table \ref{big_table}, we demonstrate how the OT setup compares with vowel and alignment distances.
We find that the method has slightly worse classification performance than alignment, but higher than pure vowel distances.
In general, optimal transport with articulatory features had higher correlation with other phonetic distance measures than GenAID, with both GenAID and WavLM features having higher correlation with speaker embedding distances.
As mentioned in the introduction, we expected GenAID to have a more holistic definition of accent due to being trained to maximise classification accuracy, and these results would seem to support that notion.
OT with the WavLM features had the highest correlation with distances from the speaker embedding across all methods, but still had a high classification accuracy at this broad level.
This reconfirms the idea that speaker and accent identity are not completely unrelated.

\begin{figure}[h]
    \centering
    \includegraphics[width=\linewidth]{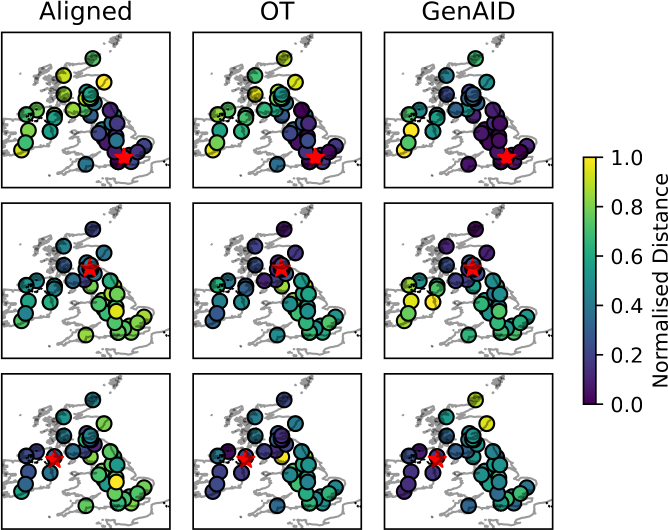}
    \caption{Normalised distances between the target speaker (red star) and all other speakers according to alignment and OT with articulatory features (Left, Middle) and GenAID embedding distance.}
    \label{speaker_maps}
\end{figure}

In Figure \ref{speaker_maps}, we plot each speaker according to the regional information given in VCTK, along with their normalised distance from the target speaker (red star) for distances defined by alignment/OT with the Art+VVN feature set and GenAID.
It seems that the alignment and OT methods with articulatory features seem to encode fairly similar information about geographically close speakers.
For articulatory distances, the boundaries between Scottish and Irish speakers are not as clear as for English (England), which reflects similar results from vowel distance literature \cite{ferragne2010vowel}.
However, for GenAID, the boundaries between all countries are quite distinct, with a slight similarity between English (England) and Scottish speakers (which may be a result of data pollution in CommonAccent \cite{xinyuan2025scalable}).

\subsection{Interpretability}

\begin{figure}[!htp]
    \centering
    \includegraphics[width=.9\linewidth]{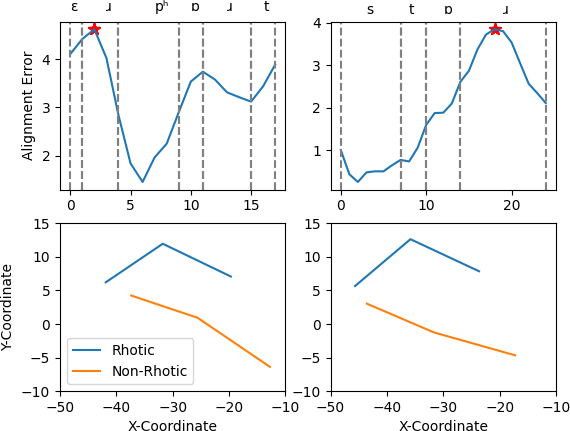}
    \caption{Tongue position for Rhotic and Non-Rhotic speakers at maximum alignment error in the word `airport' (Left) and `store' (Right).}
    \label{alignment_tongue_rhoticity}
\end{figure}

Achieving high classification accuracy is positive, but our real aim is to improve the interpretability of these methods so we can identify specific differences in pronunciation between speakers.
To that end, we look to see whether we can identify rhoticity (or lack thereof) in both the alignment and optimal transport settings.
Rhoticity is common in Irish accents but it is not present in most accents in England. 
We found that the alignment error would often peak in these areas when compared with English speakers, as is shown in Figure \ref{alignment_tongue_rhoticity} for two speakers, one from Dublin (labelled as `Rhotic') and one from Stockton-on-Tees (labelled `Non-Rhotic').

\begin{figure}[h]
    \centering
    \includegraphics[width=\linewidth]{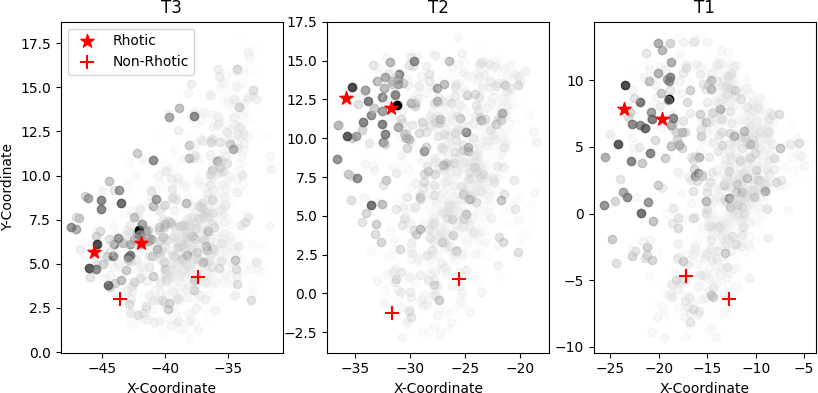}
    \caption{Rhotic (star) and non-rhotic (cross) tongue positions for `airport' and `store' compared with articulatory clusters coloured by intensity according to OT distance.}
    \label{OT_tongue_positions}
\end{figure}

The tongue position is retracted for the /\textturnr/ sound with the rhotic speaker and lowered for the vowel with the non-rhotic speaker.
We also wanted to see whether this appeared in the OT results, so in Figure \ref{OT_tongue_positions}, we plot the tongue positions for the rhotic and non-rhotic speakers from Figure \ref{alignment_tongue_rhoticity} alongside the rhotic speaker's clusters that were used to perform OT.
We adjust the intensity value of each cluster in the figure according to the distance that the cluster moved during OT, so darker clusters moved further.
The tongue positions for the rhotic speaker are located in the same area as those clusters which had to move further during OT.
Whilst we cannot determine that this position exactly constitutes a rhotic /\textturnr/, as Irish speakers can have higher back vowels than English speakers, it at least demonstrates that OT can be used to identify consonant and vowel areas that may be different between speakers.

\subsection{Optimal Transport Ablations}

\begin{figure}[h]
    \centering
    \includegraphics[width=\linewidth]{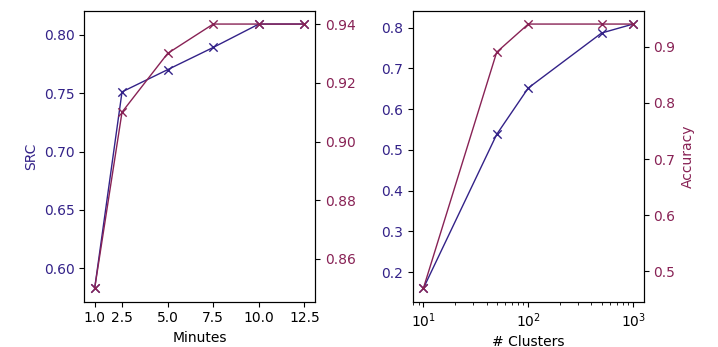}
    \caption{Effect of cluster size and number of minutes input on alignment distance SRC and classification accuracy.}
    \label{ablations}
\end{figure}

In Figure \ref{ablations}, we show that the performance of the OT method relies on the quantity of input information, both in terms of the amount of recorded speech and also the number of clusters.
We saw the maximum classification accuracy at 7.5 minutes of speech per speaker and 100 clusters, but could improve SRC with alignment distances by including more data and clusters.
It is possible that we might be able to improve both the data efficiency and interpretability of the OT method by focusing on articulatory targets instead of using all available frames.

\section{Conclusion}

In this paper, we have demonstrated that articulatory features can be used to achieve similar accent classification performance to accent-classification models through feature alignment.
Using optimal transport, we extended this to arbitrary recording types, albeit at a slightly lower classification accuracy.
We then showed that these features are far more interpretable than embeddings derived from supervised tasks, identifying rhoticity differences between two speakers in VCTK through both alignment and OT.
In future work, we'd like to see whether we can improve the data efficiency and interpretability of the OT method by clustering articulatory targets instead of clustering every available frame.
We will also look at using the distances established through alignment and OT to control an accented TTS system.
\\

\section{Generative AI Use Disclosure}

There was no use of generative AI in the production of this paper.

\bibliographystyle{IEEEtran}
\bibliography{mybib}

\end{document}